# An Online Question Answering System based on Sub-graph Searching

Shuangyong Song

Alibaba Group, Beijing 100102, China
songshuangyong@163.com

**Abstract.** Knowledge graphs (KGs) have been widely used for question answering (QA) applications, especially the entity based QA. However, searching answers from an entire large-scale knowledge graph is very time-consuming and it is hard to meet the speed need of real online QA systems. In this paper, we design a sub-graph searching mechanism to solve this problem by creating sub-graph index, and each answer generation step is restricted in the sub-graph level. We use this mechanism into a real online QA chat system, and it can bring obvious improvement on question coverage by well answering entity based questions, and it can be with a very high speed, which ensures the user experience of online QA.

## 1 Introduction

Open domain Question Answering (QA) is the task of finding answers to questions posed in natural language. With the explosion of Internet users and applications in the last decades, more and more stand-alone or subsidiary QA systems are built, and almost all of them are in the demand for very high 'question per second' (QPS), while the QPS indicates the running speed of QA system.

If too many user questions are replied with 'I don't know how to answer this question', a QA system is considered as a dull tool. Meanwhile, if a QA system takes several seconds to answer each question, this system may be too slow to be a practicable online system. Therefore, our task is to design more modules with high speed for improving the question coverage of the QA system.

Some previous works have applied sub-graph idea in KG-based QA systems, such as [1] and [2]. They try to extract answers from a question-specific sub-graph, which contains text and KB entities and relations. However, this sub-graph creation step is in the online processing part, which still slashes the online QA speed. In our work, we propose a framework, which uses a very simple offline 'frequently asked entity' oriented sub-graph extraction method to solve this problem, and online QA is just based on a Lucene index based sub-graph searching and the answer generation is restricted in the sub-graph level.

## 2 Proposed Framework

### 2.1 Sub-graph Extraction

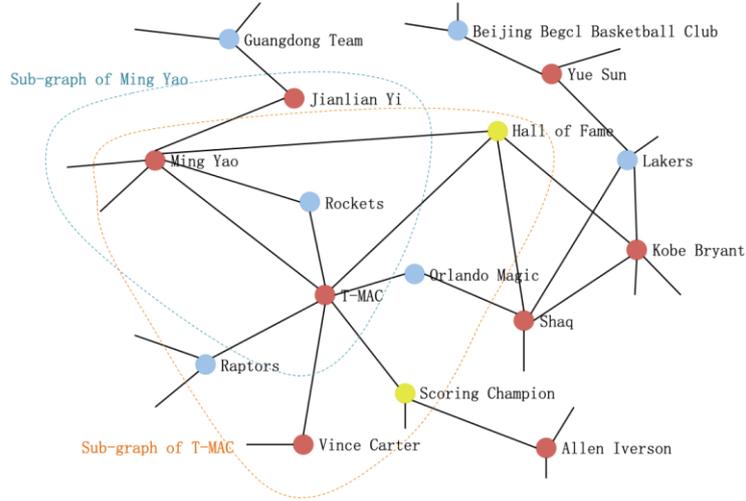

**Fig. 1.** An illustration of offline sub-graph extraction.

A hybrid method with pattern matching and sequence labeling is used to extract entities from a large amount of historical user question logs. For keeping the Lucene index with moderate scale, we just index the high-frequent entities. And for high-frequent entities, we extract sub-graphs from complete knowledge graph, and differing from some related work, which do this step in real time [1, 2].

In the sub-graph step, for each high-frequent entity we just keep all the entities with direct link as the components of its sub-graph. Figure 1 shows the illustration of offline sub-graph extraction with examples.

### 2.2 Sub-graph Based QA

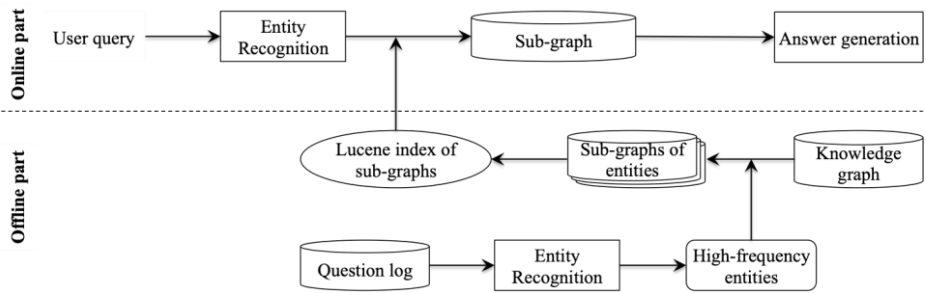

**Fig. 2.** The framework of sub-graph based QA.

Figure 2 gives the framework of sub-graph based QA. In the offline part, we create Lucene index of high-frequency entity sub-graphs. In the online part, with a user query, we recognize entities and recall sub-graphs from the Lucene index.

In the answer generation step, we use pattern-based method instead of common generation based model, such as RNN, because that may make online QA slow and the effort of sub-graph based acceleration will be offset.

### 2.3   An Entire QA System

Our online QA system, AliMe, is presented by Alibaba in 2015 and has provided services for billions of users and now on average with ten million of users access per day. Text matching models are effective and common in answering frequently asked questions (FAQ), and we choose a triple convolutional neural network model (TCNN) [4] as the matching model in AliMe, while the FAQs are collected with the clustering model proposed in [5]. However, the matching model may performs not well on entity related QA, so the above sub-graph searching based entity related QA is taken as a supplemental ability of the main text matching based QA module.

In particular, for user questions which contain more than one entity, we directly refer a method proposed in [3] to answer this kind of questions, since them just make up a small percentage of all user questions, so we don't treat them with the sub-graph searching method.

## 3    Experiments

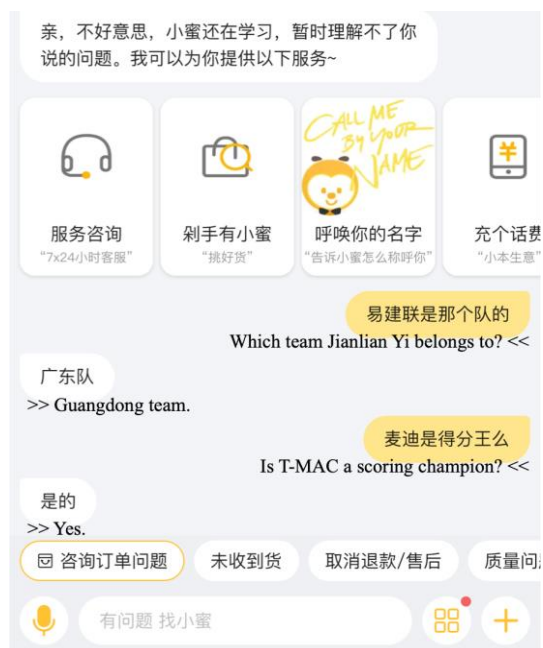

**Fig. 3.** An example of sub-graph based QA in AliMe system.

### 3.1 Dataset

Question log: we collect anonymous online user question log from April 1, 2020 to April 31, 2020. This dataset contains 95.4 million user questions and with merging duplicate ones we can obtain 28.6 million diverse user questions.

High frequent entities: 33,242 high frequent entities, each of which the occurrence frequency is more than 10, are indexed with Lucene and the sub-graphs of them are correspondingly extracted from Wikipedia..

### 3.2 Experimental Results

The main purpose of the proposed framework is to increase the coverage of *AliMe Chat* and reduce the 'no-answer' situations, with fast enough speed. With the real online testing, the coverage of *AliMe Chat* in the whole *Alime Assist* has been increased from 4.18% to 4.98%, which realizes a 19.14% increase. Meanwhile, it can reach the speed of 18.2ms per QA, which is fast enough to meet online QA needs.

In Fig. 3, we show an example of online results of the proposed framework. The first user question contains a subject and a predicate, and our task is to detect the right object with the subject related sub-graph. The second question needs the QA system to make a judgment, which is also easy with the subject related sub-graph.

## 4 Future Works

In our question answering system, open-domain chat is just an accessory module, while the product related pre-sales QA and after-sales service QA are real core modules. Therefore, product knowledge graphs and service knowledge graphs will be considered in the main part of our QA system.